\documentclass[runningheads]{llncs}
\usepackage{mathtools}
\usepackage{amsmath}
\usepackage{amssymb}

\usepackage[font=small,skip=0pt]{caption}
\begin{document}
\title{Dual Conditioned Diffusion Models for Out-Of-Distribution Detection: Application to Fetal Ultrasound Videos}
\titlerunning{Dual Conditioned Diffusion Models for OOD detection}
%
%
\authorrunning{Mishra et al.}
%
%
\author{Divyanshu Mishra\inst{1}, He Zhao\inst{1}, Pramit Saha\inst{1}, Aris T. Papageorghiou\inst{2}, J.Alison Noble\inst{1}}
\institute{Institute of Biomedical Engineering, University of Oxford
\and Nuffield Department of Women's and Reproductive Health, University of Oxford}


\maketitle              
%
\begin{abstract}
Out-of-distribution (OOD) detection is essential to improve the reliability of  machine learning models by detecting samples that do not belong to the training distribution. 
Detecting OOD samples effectively in certain tasks can pose a challenge because of the substantial heterogeneity within the in-distribution (ID),
and the high structural similarity between ID and OOD classes. For instance, when detecting heart views in fetal ultrasound videos there is a high structural similarity between the heart and other anatomies such as the abdomen, and large in-distribution variance as a heart has 5 distinct views and structural variations within each view.
To detect OOD samples in this context, the resulting model should generalise to the intra-anatomy variations while rejecting  similar OOD samples.
In this paper, we introduce dual-conditioned diffusion models (DCDM) where we condition the model on in-distribution class information and latent features of the input image for reconstruction-based OOD detection. This constrains the generative manifold of the  model to generate  images structurally and semantically 
similar to those within the in-distribution. The proposed model outperforms reference methods with a 12\% improvement in accuracy, 22\% higher precision, and an 8\% better F1 score.
\end{abstract}
\section{Introduction}
Existing out-of-distribution (OOD) detection methods work well when the in-distribution (ID) classes have low heterogeneity (low variance) but fail when in-distribution classes have high heterogeneity~\cite{sabokrou2018adversarially} or high spatial similarity between ID and OOD classes~\cite{fort2021exploring}. 
Fetal ultrasound (US) anatomy detection is one such application where both the challenges co-exist. 




In this paper, we propose a Dual-Conditioned Diffusion Model~(DCDM) to detect OOD samples when in-distribution data has high variance and test the performance by detecting heart views in fetal US videos as an example application.
Specifically, an Ultrasound (US) typically comprises 13 anatomies and their views. However,analysis models are usually developed for anatomy-specific tasks. Hence, to separate heart views from other 12 anatomies (head, abdomen, femur etc) we develop an OOD detection algorithm.
Our in-distribution data comprises five structurally different heart views captured across different cardiac cycles of a beating heart during obstetric US scanning.
We develop a diffusion-based model for  reconstruction-based OOD detection, which extends~\cite{ho2020denoising} with a novel dual conditioning mechanism that alleviates the influence of high inter- and intra-class variation within different classes by leveraging in-distribution class conditioning (IDCC) and latent image feature conditioning (LIFC).
These conditioning mechanisms allow our model to generate images similar to the input image for in-distribution data.
The primary contributions of our paper are summarized as follows:
1) We introduce a novel conditioned diffusion model for OOD detection and demonstrate that the dual conditioning mechanism is  effective in tackling challenging scenarios where in-distribution data comprises multiple heterogeneous classes and there is a high spatial similarity between ID and OOD classes.
2) Two original conditions are proposed for the diffusion model, which are in-distribution class conditioning (IDCC) and latent image feature conditioning (LIFC). IDCC is proposed to handle high inter-class variance within in-distribution classes and high spatial similarity between ID and OOD classes.
LIFC is introduced to counter the intra-class variance within each class.
3) We demonstrate in our experiments that DCDM can detect and separate heart views from other anatomies in fetal ultrasound videos without needing any labelled data for OOD classes.
Extensive experiments and ablations demonstrate superior performance over existing OOD detection methods. Our approach is not fetal ultrasound specific and could be applied to other OOD applications.

\section{Related Work}
OOD detection \cite{yang2021generalized}  involves identifying samples that do not belong to the training distribution. 
Such models can be categorized  into: (a) unsupervised OOD detection~\cite{sabokrou2018adversarially}\ 
and (b) supervised OOD detection. 
 \cite{zhou2021step,devries2018learning,guenais2020bacoun}.
Unsupervised OOD detection methods can again be
divided into two main categories:  (i) likelihood-based approaches~\cite{hendrycks2016baseline,ren2019likelihood,xu2018unsupervised}, and (ii) reconstruction-based \cite{chen2018unsupervised,zhou2022rethinking,schlegl2017unsupervised}.
Likelihood-based approaches suffer from several issues, including assigning higher likelihood to OOD samples 
 \cite{nalisnick2018deep,choi2018waic}, susceptibility to adversarial attacks~\cite{fort2022adversarial}, and calibration issues~\cite{wald2021calibration}.
Current reconstruction-based approaches are sensitive to dimensions of the bottleneck  layer and require rigorous tuning specific to the dataset and task \cite{graham2022denoising}. Additionally, models trained using a generator-discriminator architecture and optimizing adversarial losses can be highly unstable and challenging to train~\cite{arjovsky2017towards,bau2019seeing}.
Finally, reconstruction-based methods often rely on highly compressed latent representations, which can lead to loss of important low-level detail. This can be problematic when discriminating between classes with high spatial similarity. Recently, diffusion models have been introduced to address these limitations on tasks such as image synthesis \cite{dhariwal2021diffusion}, and OOD detection \cite{graham2022denoising}.

Denoising Diffusion Probabilistic Models (DDPMs) \cite{ho2020denoising} are generative models that work by gradually adding noise to an input image through a forward diffusion process followed by gradually removing noise using a trained neural network in the backward diffusion process 
\cite{yang2022diffusion}. 
To guide the generative process of a diffusion model (DM), previous work \cite{meng2021sdedit,saharia2022palette,rombach2022high} condition the DDPMs on task-specific conditioning. In image-to-image translation tasks like super-resolution, colourization, \textit{etc.},  previous papers \cite{saharia2022palette} condition the model by concatenating a resized or grayscale version of the input image to the noised image. This concatenation is unsuitable for reconstruction-based OOD detection as the model will generate similar images for ID and OOD samples.
In the context of OOD detection using DMs, previous works \cite{graham2022denoising} have trained unconditional DDPMs and, during inference, sampled using a Pseudo Linear Multi Step (PLMS) \cite{liu2022pseudo} sampler  for varying noise levels. However, their approach generates 5500 samples to detect OOD samples for each input image which is time-consuming and impractical for settings where shorter inference times are needed.
AnoDDPM \cite{wyatt2022anoddpm} utilises simplex noise rather than Gaussian noise to corrupt the image (t=250 rather than t=1000) for anomaly detection. However, this approach requires data specific tuning,and is outperformed by \cite{graham2022denoising}.
\begin{figure}
    \centering
    
    \includegraphics[width=100mm]{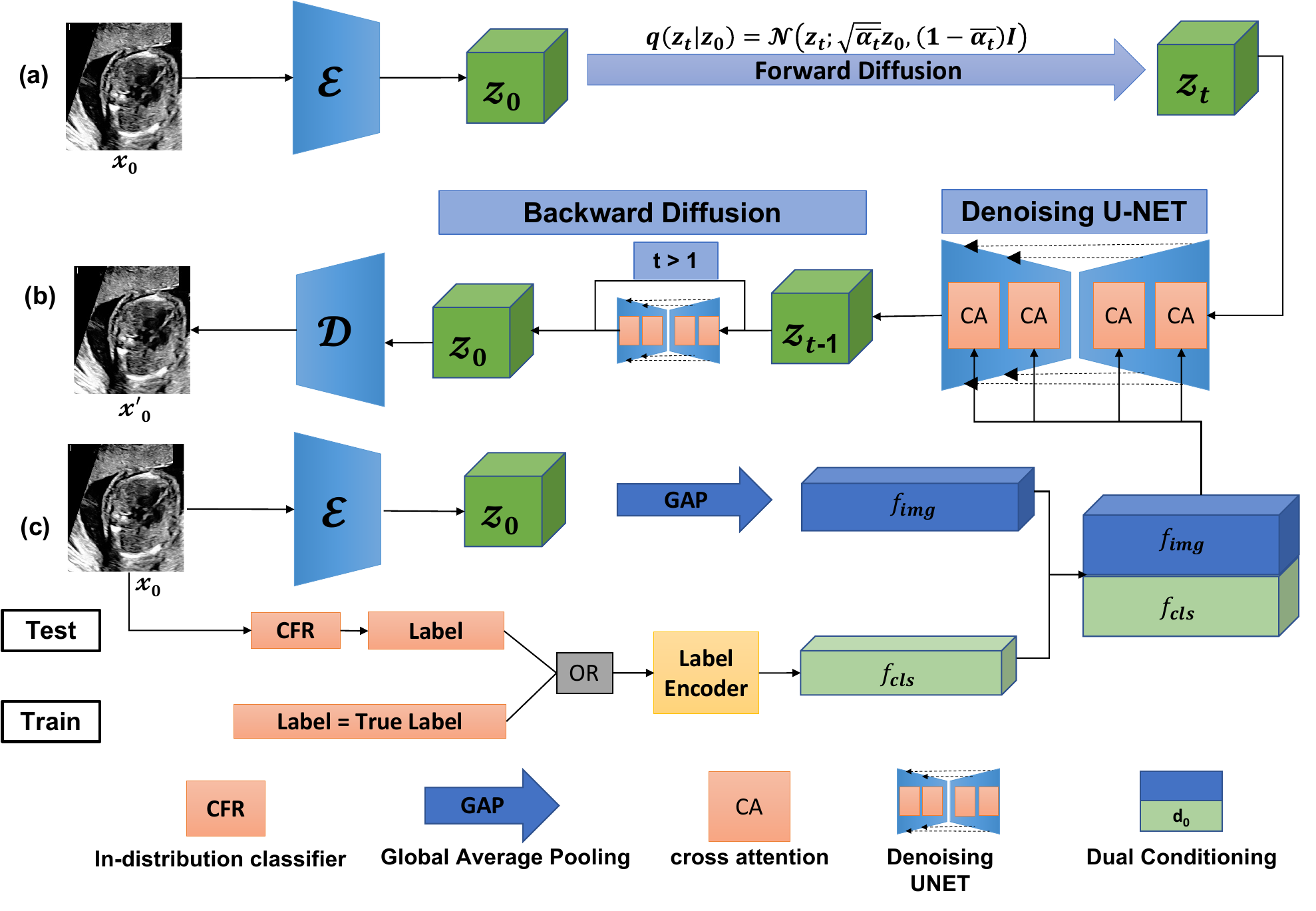}
    \caption{DCDM architecture where $\boldsymbol{(a)}$  the input image $x_0$ is mapped to the latent vector $z_0$ using a pretrained encoder $\mathcal{E}$ and forward diffusion is applied, $\boldsymbol{(b)}$ the backward diffusion process denoises the latent vector $z_t$ and the final denoised latent vector $z_0$ is mapped to pixel space by the decoder $\mathcal{D}$ $\boldsymbol{(c)}$ the dual-conditioning mechanism. We obtain ${f_{img}}$ by passing the input image $x_0$ through the encoder $\mathcal{E}$. ${f_{cls}}$ is obtained using the true label during training or predicted class label during testing.}
    \label{figure1}
\end{figure}

\section{Methods}


\subsection{Dual Conditioned Diffusion Models}
Diffusion models are generative models that rely on two Markov processes known as forward and backward diffusion~\cite{ho2020denoising}.
To improve efficiency during training and inference, forward and backward diffusion  is applied to the latent space~\cite{rombach2022high}. Autoencoder (AE = $\mathcal{E}$ + $\mathcal{D}$) is pretrained separately on ID heart data and can successfully reconstruct the input heart images (SSIM=0.956). The latent variable ${z_0}$ is obtained by passing an input image ${x_0}$ through a pretrained encoder $\mathcal{E}$.
Given the latent vector  ${z_0}$ and a  fixed variance schedule~\cite{ho2020denoising} $\{\beta_t \in (0, 1)\}_{t=1}^T$, the forward diffusion process,  defined by Eqn.~\ref{eq2}, gradually adds Gaussian noise to ${z_0}$ to give a noised latent vector ${z_t}$ where ${\alpha_t = 1 - \beta_t}$ and 
{$\bar{\alpha}_t = \prod_{i=1}^t \alpha_i$}:
\begin{equation}
\label{eq2}
q(\boldsymbol{z}_t \vert \boldsymbol{z}_0) = \mathcal{N}(\boldsymbol{z}_t; \sqrt{\bar{\alpha}_t} \boldsymbol{z}_0,(1 - \bar{\alpha}_t)\mathbf{I})
\end{equation}
In backward diffusion, we aim to reverse the forward diffusion process and predict {$z_{t-1}$} given {$z_t$}. To predict {(${z}_{t-1} \vert {z}_t$)},we train  a denoising U-Net~\cite{ho2020denoising} denoted as {$\boldsymbol{\epsilon}_\theta \left(z_t, t,d_0\right)$} that takes the current timestep $t$, noised latent vector {$z_t$} and the dual conditioning embedding vector {$d_0$} as input and predicts the noise at timestep $t$ as shown in Eqn. \ref{eq3.1}. 
\begin{equation}
\label{eq3.1}
{\boldsymbol{z}_{t-1}  = \mathcal{N}(\boldsymbol{z}_{t-1}; \frac{1}{\sqrt{\alpha_t}} \Big( \boldsymbol{z}_t - \frac{1 - \alpha_t}{\sqrt{1 - \bar{\alpha}_t}} \boldsymbol{\epsilon}_\theta(\boldsymbol{z}_t, t,d_0) \Big), {(1 - \bar{\alpha}_t)\mathbf{I}})}
\end{equation}

\noindent The dual embedding vector $d_0$ is obtained by combining  IDCC (${f_{cls}}$) and LIFC (${f_{img}}$) vectors, which we explain  in Section 3.2.
The output ${z_{t-1}}$ is again input to ${\epsilon_\theta}$. This process is repeated until $z_0$ is obtained.





The final model optimisation objective is given by Eqn. \ref{eq4} where $\epsilon$ is the original noise added during the forward diffusion process.
\begin{equation}
   \label{eq4}
\mathcal{L}_{DCDM}:=\mathbb{E}_{\mathcal{E}(x), \epsilon \sim \mathcal{N}(0,1), t}\left[\left\|\epsilon-\epsilon_\theta\left(z_t, t,d_0\right)\right\|_2^2\right]
\end{equation}
Once we obtain {$z_0$} from the backward diffusion process, it is passed on to the decoder $\mathcal{D}$ and mapped back to the pixel space to give generated image {$x'_0$}. 

\subsection{Dual Conditioning Mechanism}
Image features and in-distribution class information are utilised in our proposed dual conditioning mechanism. This guides the DCDM to generate images that are spatially and semantically similar to the input image for in-distribution samples and dissimilar for OOD samples.

\noindent \textbf{Latent Image Feature Conditioning (LIFC):}
The image conditioning dictates the desired appearance of  generated images in terms of shape and texture.
In our model, we use the features extracted by a pretrained encoder for conditioning.
Empirically, we use the same encoder $\mathcal{E}$ as our feature extractor to obtain latent feature vector $z_0$ as shown in Fig.~\ref{figure1}. Specifically, the input image of dimension $224 \times 224 \times 3$ is passed through the encoder $\mathcal{E}$ and a feature map with the size of $7 \times 7 \times 128$ is obtained which is followed by global average pooling (GAP) resulting in  a feature vector ({$f_{img}$}) with dimension 128.

\noindent\textbf{In-Distribution Class Conditioning (IDCC):}
Given an in-distribution dataset comprising $n$ heterogeneous classes, conditioning the model only on image-level features is insufficient. Therefore we introduce an in-distribution class conditioning (IDCC) that informs the DCDM of the class of the input image and enables it to generate samples belonging to the same class for ID.
A label encoder generates a unique class conditional embedding ($f_{cls}$) of dimension 128 for each class label.
The class label is assigned based on the ground truth label  during the training phase and to the classifier's prediction during inference, as depicted in Fig.~\ref{figure1}. 
In practice, we train a CNN classifier, freeze its weight and use it as our in-distribution classifier (CFR), as discussed in section 3.3.

\noindent\textbf{Cross Attention Guidance:}
To integrate the dual-conditioning guidance into the diffusion model, we use a cross-attention \cite{vaswani2017attention} mechanism inside the denoising U-Net rather than just concatenation \cite{saharia2022palette} as it is more effective \cite{hertz2022prompt,rebain2022attention,margatina2019attention} and allows  condition diffusion models on various input modalities \cite{rombach2022high}.
Our LIFC and IDCC are first concatenated to give a feature vector with a dimension of 256. This acts as a side input to each UNet block. The features from the UNet block and the conditional features are fused by cross-attention and serve as input to the following UNet block as shown in Fig.~\ref{figure1}.For more details,regarding cross-attention block refer to Rombach \emph{et al} \cite{rombach2022high}.




\subsection{In-Distribution Classifier}
The in-distribution classifier (CFR) serves two main functions. First, it provides labels for the class conditioning during inference; second, it is utilized as a feature extractor for calculating the OOD score.

\noindent\textbf{Inference Class Guidance.}
IDCC requires in-distribution class information to generate the class conditional embedding. However, class information is only available during training. To obtain class information during inference, we separately train a ConvNext CNN based classifier (accuracy = 88\%) on the in-distribution data and use its predictions as the class information.
During inference, the input image {$x_0$} is passed through the classifier, and the predicted label is used to generate the class embedding by feeding to the label encoder as shown in Fig. \ref{figure1}. Moreover, as the classifier is only trained on in-distribution data, it classifies an OOD sample to an in-distribution class. The classifier's prediction is utilised by the DCDM and it tries to generate an image belonging to in-distribution class for the OOD samples. This reduces the structural and semantic similarity between the input and the generated image, as demonstrated by our qualitative results (Fig. \ref{fig2}).

\noindent\textbf{Feature-Based OOD Detection}
To evaluate the performance of the DCDM, the cosine similarity between features of the input image $x_0$ and the generated image $x'_0$ from the in-distribution classifier is calculated and is referred as an \text{OOD score} where $f_0$ and $f'_0$ are the features of $x_0$ and $x'_0$, respectively:

\begin{equation}
\label{eq6}
\text{OOD~score} = ~\\sim(f_0,f'_0 )~ =   \dfrac {f_0 \cdot f'_0} {\left\| f_0\right\| _{2}\left\| f'_0\right\| _{2}},
\end{equation}
An input image $x_0$ is classified as in-distribution (ID) or OOD based on Eqn. \ref{eq8} where $\tau$ is a pre-defined threshold and $y_{pred}$ is the prediction of our feature-based OOD detection algorithm.
\begin{equation}
  \label{eq8}
y_{pred} = \begin{cases}
0 (ID) & if \hspace{1mm} \text{OOD~score}  >\tau \\ 
1~\text{(OOD)} & otherwise
\end{cases}
\end{equation}

\section{Experiments and Results}
\subsubsection{Dataset and Implementation}
For our experiments, we utilized a fetal ultrasound dataset of 359 subject videos that were collected as part of the PULSE  project \cite{drukker2021transforming}. The in-distribution dataset consisted of 5 standard heart views (3VT, 3VV, LVOT, RVOT, and 4CH), while the out-of-distribution dataset comprised of three non-heart anatomies - fetal head, abdomen, and femur. The original images were of size $1008 \times 784$ pixels and were resized to $224 \times 224$ pixels.  

To train the models, we randomly sampled 5000 fetal heart images and used 500 images for evaluating image generation performance. To test the performance of our final model and compare it with other methods, we used an held-out dataset of 7471 images, comprising 4309 images of different heart views and 3162 images (about 1000 for each anatomy) of out-of-distribution classes.
Further details about the dataset are given in \textbf{Supp. Fig. 2 and 3}.

All models were trained using PyTorch version 1.12 with a Tesla V100 32 GB GPU. During training, we used T=1000 for noising the input image and a linearly increasing noise schedule that varied from 0.0015 to 0.0195. To generate samples from our trained model, we used DDIM \cite{song2020denoising} sampling with T=100. 
All baseline models were trained and evaluated using the original implementation.
\subsection{Results}

We evaluated the performance of the dual-conditioned diffusion models (DCDMs) for OOD detection by comparing them with two current state-of-the-art unsupervised reconstruction-based approaches and one likelihood-based approach.
The first baseline is Deep-MCDD~\cite{lee2020multi}, a likelihood-based OOD detection method that proposes a  Gaussian discriminant-based objective to learn class conditional distributions. The second baseline is 
ALOCC~\cite{sabokrou2018adversarially} a GAN-based model that uses the confidence of the discriminator on reconstructed samples to detect OOD samples. The third baseline is the method of Graham \emph{et al.} \cite{graham2022denoising}, where they use  DDPM~\cite{ho2020denoising} to generate multiple images at varying noise levels for each input. They then compute the MSE and LPIPS metrics for each image compared to the input, convert them to Z-scores, and finally average them to obtain the OOD score.

\noindent\textbf{Quantitative Results}
 The performance of the DCDM, along with comparisons with the other approaches, are shown in Table 1. The GAN-based method ALOCC \cite{sabokrou2018adversarially} has the lowest AUC of 57.22\%, which is improved to 63.86\% by the method of Graham \emph{et al.} and further improved to 64.58\% by likelihood-based Deep-MCDD. DCDM outperforms all the reference methods by 20\%, 14\% and 13\%, respectively and has an AUC of 77.60\%.
 High precision is essential for OOD detection as this can reduce false positives and increase trust in the model. DCDM exhibits a precision that is 22\% higher than the reference methods while still having an 8\% improvement in F1-Score.

\begin{figure}[t]
    \centering  \includegraphics[width=80mm]{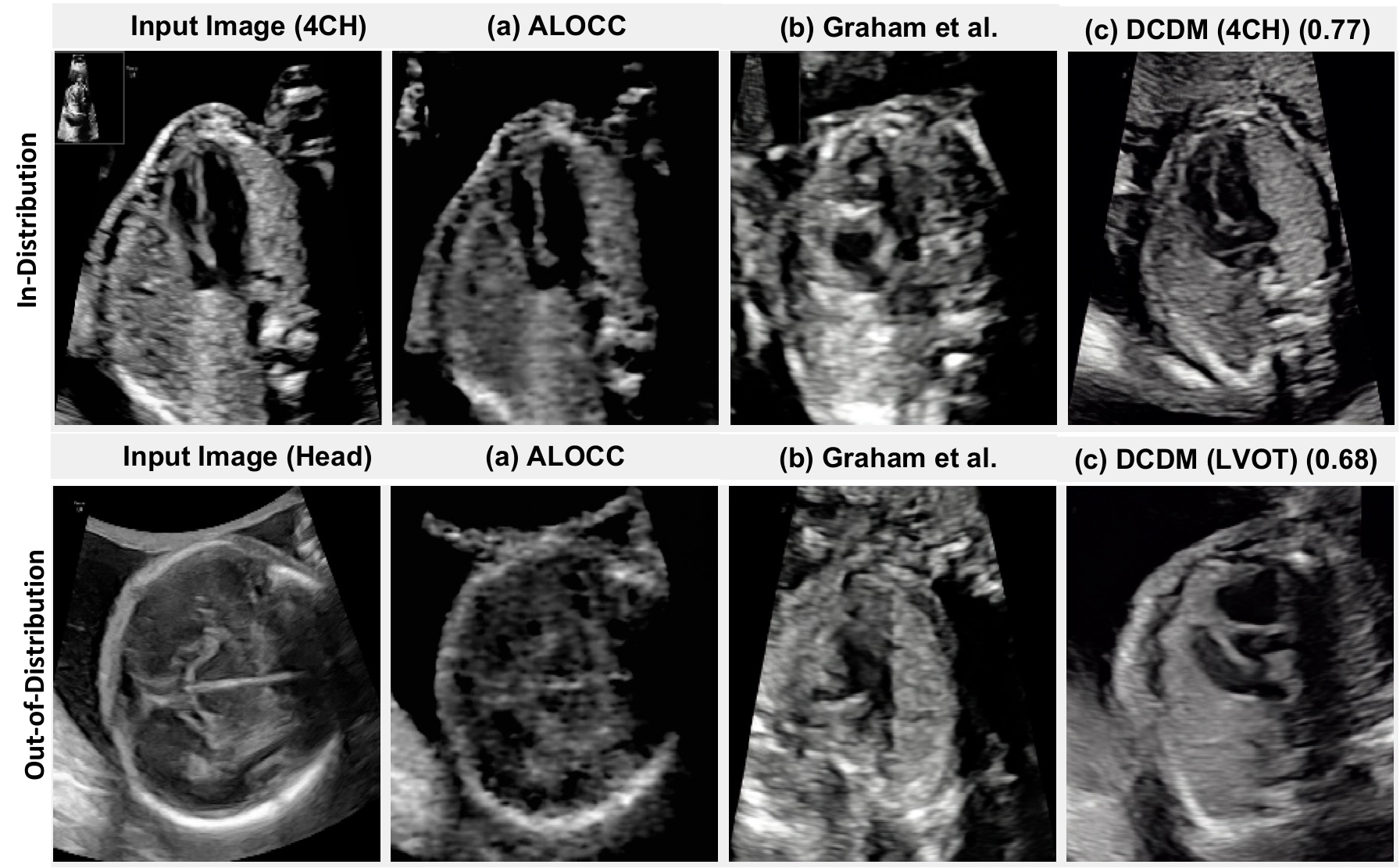}
    \caption{Qualitative comparison of our method with $\boldsymbol{(a)}$ ALOCC generates similar images to the input for ID and OOD samples $\boldsymbol{(b)}$ Graham \emph{et al.} generates any random heart view for a given input image $\boldsymbol{(c)}$ Our model generates images that are similar to the input image for ID and dissimilar for OOD samples. Classes predicted by CFR and the OOD score ($\tau$ = 0.73) are mentioned in brackets. }
    \label{fig2}
\end{figure}

\noindent\textbf{Qualitative Results}
Qualitative results are shown in Fig.~\ref{fig2}.
Visual comparisons show ALOCC generates images  structurally similar to input images for in-distribution and OOD samples. This makes it harder for the ALOCC model to detect OOD samples.
The model of Graham~\emph{et al.} generates any random heart view for a given image as a DDPM is unconditional, and our in-distribution data contains multiple heart views. For example, given a 4CH view as input, the model generates an entirely different heart view. However, unlike ALOCC, the Graham~\emph{et al.} model generates heart views for OOD samples, improving OOD detection performance. DCDM generates images with high spatial similarity to the input image and belonging to the same heart view for ID samples while structurally diverse heart views for OOD samples. In Fig \ref{fig2} (c) for OOD sample, even-though the confidence is high (0.68), the gap between ID and OOD classes is wide enough to separate the two.Additional qualitative results can be observed in \textbf{Supp. Fig. 4.}

\begin{table}[t]
\centering

\caption{Quantitative comparison of our model (DCDM) with reference methods}
\label{tab1}
\scalebox{0.85}{
\begin{tabular}{|c|c|c|c|c|}
\hline
\textbf{Method} & \textbf{AUC(\%)} & \textbf{F1-Score(\%)} & \textbf{Accuracy(\%)} & \textbf{Precision(\%)} 
\\ \hline
Deep-MCDD \cite{lee2020multi}          & 64.58            & 66.23                 & 60.41                 & 51.82                  
\\ \hline
ALOCC \cite{sabokrou2018adversarially}          & 57.22            & 59.34                 & 52.28                 & 45.63               
\\ \hline
Graham et al. \cite{graham2022denoising}   & 63.86            & 63.55                 & 60.15                 & 50.89                 
\\ \hline
DCDM(Ours)      & \textbf{77.60}   & \textbf{74.29}        & \textbf{77.95}        & \textbf{73.34}         
\\ \hline
\end{tabular}}

\end{table}
\subsection{Ablation Study}

Ablation experiments were performed to study the impact of various conditioning mechanisms on the model performance both qualitatively and quantitatively.
When analyzed quantitatively, as shown in Table  \ref{tab2}, the unconditional model has the lowest AUC of 69.61\%. Incorporating the IDCC guidance or LIFC separately, improves performance with an AUC of 75.27\% and 77.40\%, respectively.
The best results are achieved when both mechanisms are used (DCDM), resulting in an 11\% improvement in the AUC score relative to the unconditional model. 
Although there is a small margin of performance improvement between the combined model (DCDM) and the LIFC model in terms of AUC, the precision improves by 3\%, demonstrating the combined model is more precise and hence the best model for OOD detection.

\begin{figure}[t]
    \centering    \includegraphics[width=100mm]{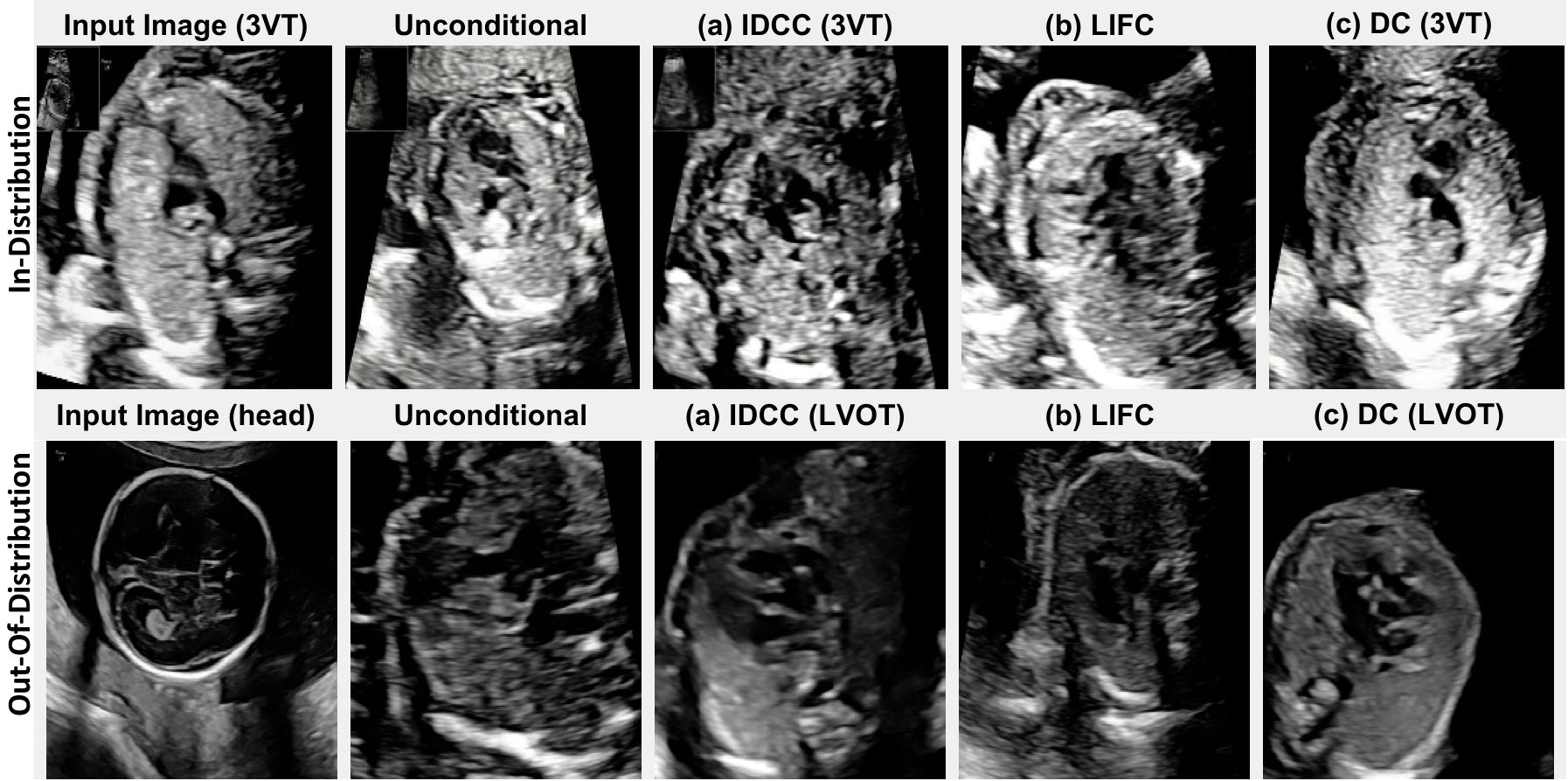}
    \caption{Qualitative ablation study showing the effect of \textbf{(a)} IDCC, \textbf{(b)} LIFC  and, \textbf{(c)} DC on generative results of DM. Brackets in IDCC, DC show labels predicted by CFR.}
    \label{fig3}
    \vspace{-3mm}
\end{figure}

As shown in Fig. \ref{fig3},  the unconditional diffusion model generates a random heart view for a given input for both in-distribution and OOD samples. 
The IDCC guides the model to generate a heart view according to the in-distribution classifier (CFR) prediction which leads to the generation of similar samples for in-distribution input while dissimilar samples for OOD input. 
On the other hand, LIFC generates an image with similar spatial information. However, heart views are still generated for OOD samples as the model was only trained on them. When dual-conditioning (DC) is used, the model generates images that are closer aligned to the input image for in-distribution input and high-fidelity heart views for OOD than those generated by a model conditioned on either IDCC or LIFC alone. \textbf{Supp. Fig. 1} presents further qualitative ablations.
\begin{table}[t]
\centering
\caption{Ablation study of different conditioning mechanisms of DCDM.
}
\label{tab2}
\scalebox{0.8}{
\begin{tabular}{|c|c|c|c|}
\hline
\textbf{Method}                   & \multicolumn{1}{l|}{\textbf{Accuracy (\%)}} & \multicolumn{1}{l|}{\textbf{Precision (\%)}} & \multicolumn{1}{l|}{\textbf{AUC (\%)}} \\ \hline
Unconditional                     & 68.16                                          & 58.44                                           & 69.61                                     \\ \hline
In-Distribution Class Conditioning   & 74.39                                          & 66.12                                           & 75.27                                     \\ \hline
Latent Image Feature Conditioning & 77.02                                          & 70.02                                           & 77.40                                     \\ \hline
Dual Conditioning                 & \textbf{77.95}                                          & \textbf{73.34}                                           & \textbf{77.60}                                     \\ \hline
\end{tabular}}
\end{table}

\section{Conclusion}

We introduce novel dual-conditioned diffusion model for OOD detection in fetal ultrasound videos and demonstrate how the proposed dual-conditioning mechanisms can manipulate the generative space of a diffusion model. Specifically, we show how our dual-conditioning mechanism can tackle scenarios where the in-distribution data has high inter- (using IDCC) and intra- (using LIFC) class variations and guide a diffusion model to generate similar images to the input for in-distribution input and dissimilar images for OOD input images.
Our approach does not require labelled data for OOD classes and is especially applicable  to challenging scenarios where the in-distribution data comprises more than one class and there is high similarity between the in-distribution and OOD classes.

\section{Acknowledgement}
This work was supported in part by the InnoHK-funded Hong Kong Centre for Cerebro-cardiovascular Health Engineering (COCHE) Project 2.1 (Cardiovascular risks in early life and fetal echocardiography), the UK EPSRC (Engineering and Physical Research Council) Programme Grant EP/T028572/1 (VisualAI), and a  UK EPSRC Doctoral Training Partnership award.

\
\bibliographystyle{splncs04}
\bibliography{ref}
%




\end{document}